\documentclass{article}
\usepackage{spconf,amsmath,graphicx}

\usepackage{multirow}

\usepackage{tipa}

\usepackage{kotex}

\usepackage{hyperref}

\usepackage[utf8]{inputenc}

\usepackage[T1]{fontenc}

\usepackage{hyperref}
\usepackage{cleveref}
\crefname{section}{§}{§§}
\Crefname{section}{§}{§§}

\title{Neural Grapheme-to-Phoneme Conversion with Pre-trained Grapheme Models}
\name{Lu Dong$^{1 *}$, Zhi-Qiang Guo$^1$, Chao-Hong Tan$^1$, Ya-Jun Hu$^2$, Yuan Jiang$^{1, 2}$, Zhen-Hua Ling$^1$ \thanks{$^{*}$Work done during internship at iFLYTEK Research.}}
\address{
$^1$National Engineering Laboratory for Speech and Language Information Processing, \\
University of Science and Technology of China, Hefei, P. R. China \\
$^2$iFLYTEK Research, iFLYTEK Co., Ltd., Hefei, P. R. China
\\ \small
{ \texttt{\{dl1111, gzq, chtan\}@mail.ustc.edu.cn, \{yjhu, yuanjiang\}@iflytek.com, zhling@ustc.edu.cn}
}
}

\usepackage{fancyhdr}

\begin{document}

\ninept
\maketitle

\thispagestyle{fancy}
\fancyhead{}
\lhead{}
\lfoot{\copyright 2022 IEEE. Personal use of this material is permitted. Permission from IEEE must be obtained for all other uses, in any current or future media, including reprinting/republishing this material for advertising or promotional purposes, creating new collective works, for resale or redistribution to servers or lists, or reuse of any copyrighted component of this work in other works.}
\cfoot{}
\rfoot{}
%
\begin{abstract}
Neural network models  have achieved state-of-the-art performance on 
grapheme-to-phoneme (G2P) conversion. 
However,  their performance relies on large-scale pronunciation dictionaries, which may not be available for a lot of languages.
Inspired by the success of the pre-trained language model  BERT, this paper proposes a pre-trained grapheme model called grapheme  BERT (GBERT),  which is built   by self-supervised training on a large, language-specific word list with only grapheme information. Furthermore, two approaches are developed to  incorporate GBERT into the state-of-the-art Transformer-based G2P model, i.e.,  fine-tuning GBERT or fusing GBERT into the Transformer model by attention.
Experimental results on   the  Dutch, Serbo-Croatian, Bulgarian  and Korean datasets of the SIGMORPHON 2021 G2P  task  confirm the effectiveness of our GBERT-based G2P models
under  both medium-resource and low-resource data conditions.

\end{abstract}
\begin{keywords}
 grapheme-to-phoneme conversion, pre-trained grapheme model, self-supervised training,  Transformer
\end{keywords}

%
\section{Introduction}
\label{sec:intro}


The grapheme-to-phoneme (G2P) conversion  task is  predicting the pronunciation of  words from their spellings. 
Considering that a pronunciation dictionary can never cover all possible words in a language, G2P conversion is essential for any applications that depend on the mapping relationship between the spoken and written forms of a language, such as TTS and ASR \cite{elias2021parallel, gao2021pre, masumura20_interspeech}. 
Many studies have been conducted on G2P conversion.  In early years, joint n-gram models \cite{galescu2002pronunciation}, joint sequence models \cite{bisani2008joint} and wFST \cite{novak2012wfst}
 were proposed. 
Recently, 
neural networks such as LSTM \cite{toshniwal2016jointly} and Transformer \cite{yolchuyeva2019transformer} have showed  powerful ability on G2P conversion. 
The Transformer-based models have achieved  state-of-the-art
performance in many benchmarks \cite{yolchuyeva2019transformer, gorman2020sigmorphon}. 
Some imitation learning based methods \cite{makarov2020cluzh} also achieved  comparable performance to the Transformer model.
Nevertheless, building neural G2P models usually relies on a large, language-specific pronunciation dictionary, which may not be available for a lot of languages.
One approach to address this issue is  cross-lingual modeling. An early work is a wFST-based system \cite{deri2016grapheme}. Subsequently, multilingual neural networks  \cite{peters2017massively} and pre-trained G2P models of  high-resource languages \cite{engelhart2021grapheme} showed better cross-lingual G2P modeling ability.
Another approach is utilizing multimodal data. Route et  al. \cite{route2019multimodal}  found that  additional audio supervision can help the G2P model to  learn a more optimal intermediate representation of graphemes.
However, these studies mainly focused on utilizing the data resources of other languages or other modals to improve the performance of limited-resource G2P conversion, which do not explore better grapheme representations for  G2P in an unsupervised way.

Therefore, this paper proposes a  pre-trained grapheme model named grapheme BERT (GBERT) to improve the Transformer-based G2P model. The design of GBERT is inspired by the language model BERT \cite{devlin2019bert}, which provides contextual word representations and has achieved great successes in various NLP tasks, such as machine translation \cite{zhu2019incorporating} and text summarization \cite{liu2019text}.  
Similarly, GBERT is designed to capture the  contextual relationship among the graphemes in a word,
which is essential for the G2P task since the same grapheme may have different pronunciations due to different contexts.
Following BERT, 
GBERT is a multi-layer Transformer encoder and is pre-trained by self-supervised learning on a large, language-specific word list with only grapheme  information. The pre-training task of GBERT is a masked grapheme prediction task, i.e, predicting the masked graphemes from  the seen graphemes in a word.  
Furthermore, two approaches are developed to improve the  Transformer-based G2P model with GBERT. They are fine-tuning GBERT and  fusing GBERT into the Transformer model by attention \cite{zhu2019incorporating}. 
Experiments were conducted on   the  Dutch, Serbo-Croatian, Bulgarian  and Korean datasets of the SIGMORPHON 2021 G2P task \cite{ashby2021results}. The results   show that fusing GBERT by attention can reduce the word error rate (WER) and phone error rate (PER) of G2P  under  a medium-resource condition,  while fine-tuning GBERT is effective for most languages  under a low-resource condition. It should be noted that a concurrent work T5G2P \cite{vrezavckova2021t5g2p} also mentions the pre-trained grapheme model can help the G2P conversion. The difference is that the T5G2P uses an encoder-decoder framework which captures  contextual relationship between graphemes and   autoregressive information of graphemes in the pre-training stage and only uses a directly fine-tuning method.
	\begin{figure*}[htbp]
	\setcounter{figure}{1}
		\centering
		\includegraphics[scale=0.5]{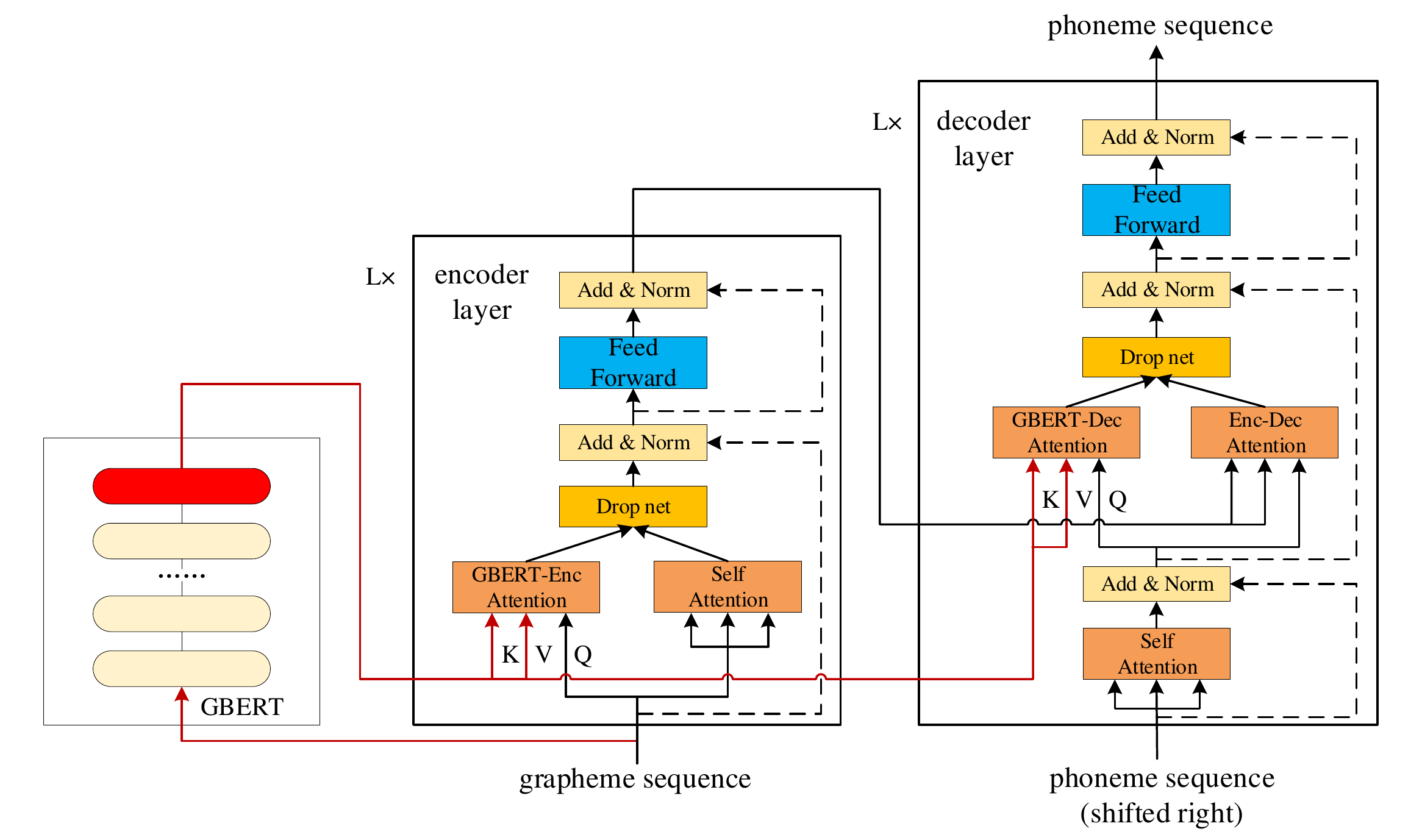}
		\caption{The architecture of GBERT-fused model, which fuses GBERT into the Transformer-based G2P model and is adapted from \cite{zhu2019incorporating}.}
		\label{fig:fig2}
	\end{figure*}
	\begin{figure}[htbp]
	\setcounter{figure}{0}
		\centering
		\includegraphics[scale=0.5]{
		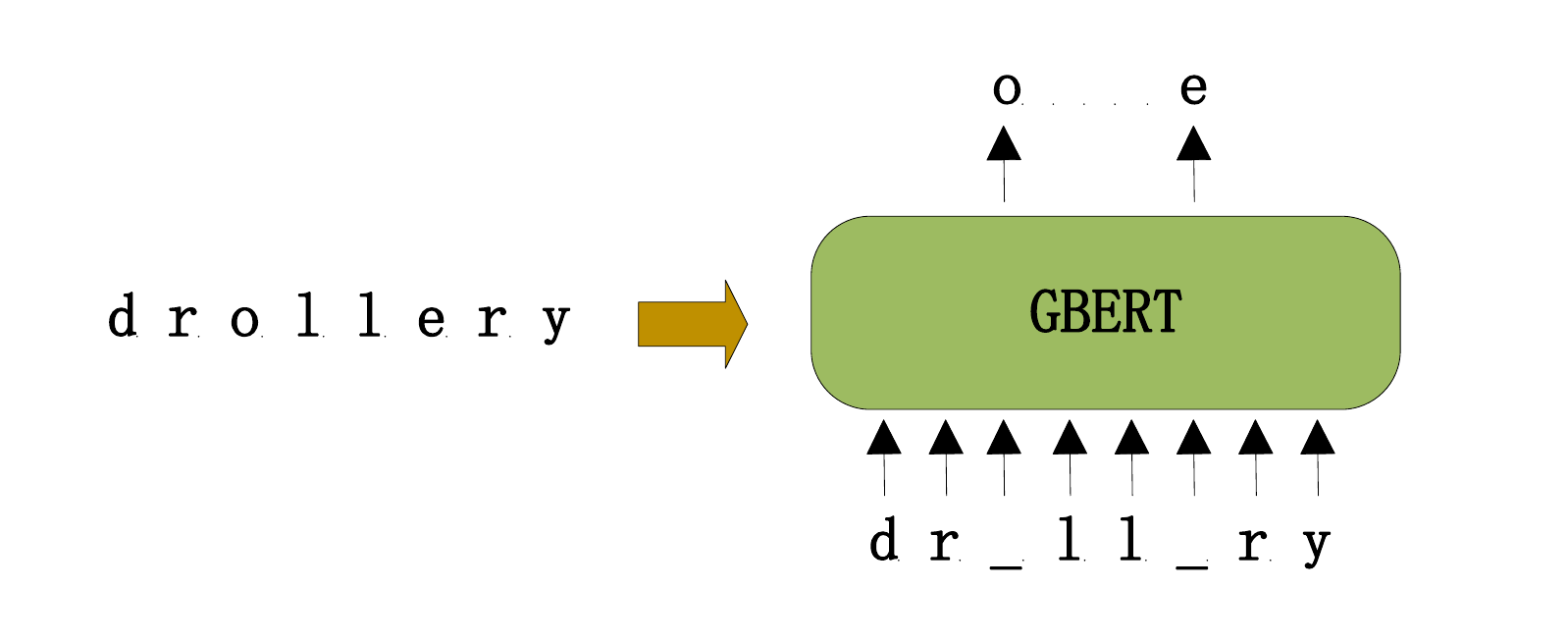}
		\caption{The masked grapheme prediction  task for pre-training GBERT. Here is an example of the English word \emph{drollery}. The ``\_'' denotes the mask token.}
			\label{fig:fig1}
	\end{figure}

\section{Proposed Method}

In this section, we first introduce our proposed pre-trained grapheme  model GBERT in Section \ref{ssec:graphemebert}. Then, we show the details of fine-tuning GBERT for G2P  in Section \ref{ssec:finetuningbert} and the details of fusing GBERT into the Transformer-based G2P model \cite{zhu2019incorporating} in Section \ref{ssec:bertattention}. 

\subsection{Grapheme BERT (GBERT)}
\label{ssec:graphemebert}
Following BERT \cite{devlin2019bert}, the model architecture of GBERT is a multi-layer bidirectional  Transformer encoder where each input token can see all input tokens. GBERT differs from BERT in that the input to GBERT is the grapheme sequence in a single word while the input  to BERT is the wordpiece sequence in a single sentence or two sentences.
The two-sentence input of  BERT is designed to cope with the  downstream tasks based on sentence pairs. However, this paper focuses on the single-word G2P task and thus we only consider the grapheme seqeunces in  single words. 

GBERT is pre-trained  using a masked grapheme prediction task, i.e., masking some percentage of the input graphemes at random and predicting those masked graphemes from the seen ones. An illustration is shown in Fig.~\ref{fig:fig1}. These mask inputs (``\_'') are replaced by mask tokens (80\% of the ratio), random graphemes (10\% of the ratio) or original graphemes (10\% of the ratio) to eliminate the  mismatch between the pre-training task and the downstream G2P task, i.e, the mask tokens will not be encountered in the G2P task.


\subsection{Fine-tuning GBERT for G2P Conversion}
\label{ssec:finetuningbert}
Fine-tuning a pre-trained language model is a typical way to apply pre-trained representations to downstream tasks. Since GBERT is a pre-trained Transformer encoder while the vanilla Transformer  includes an  encoder and a  decoder, fine-tuning GBERT means that we replace the  encoder in the vanilla Transformer with GBERT and train the new model in an end-to-end manner. Moreover, Liu and Lapata \cite{liu2019text} found that  using different learning rates for the pre-trained  encoder and the randomly initialized  decoder can lead to better convergence. We also leveraged this trick in our implementation.

\subsection{Fusing GBERT into Transformer-based G2P Model}
\label{ssec:bertattention}
In addition to fine-tuning, another approach to integrate pre-trained language models is  utilizing them as feature extractors, which may work better than the fine-tuning approach when the downstream tasks can not be easily represented by a Transformer encoder architecture \cite{devlin2019bert}. One example is the BERT-fused model \cite{zhu2019incorporating}, which adopts a multi-head attention \cite{vaswani2017attention} and a drop net to adaptively control how each encoder and decoder layer of the vanilla Transformer model interacts with the output features of BERT. Since this method worked better  than fine-tuning BERT and a na{\"i}ve feature-based method in the medium-resource machine translation task, it is adapted in this paper to fuse GBERT into the Transformer-based G2P model, aiming to achieve better results in the medium-resource G2P scenario. An illustration  is shown in Fig.~\ref{fig:fig2}, where GBERT replaces the BERT in the original BERT-fused model \cite{zhu2019incorporating}.

As shown in Fig. \ref{fig:fig2},  the GBERT-fused model consists of $L$ encoder layers and $L$ decoder layers in addition to GBERT. For the $l$-th encoder layer, its input includes the output of the $(l-1)$-th encoder layer (or the embeddings of the grapheme sequence when $l=1$) as well as the contextual grapheme representations provided by GBERT. 
For the $l$-th decoder layer, its  input includes the hidden states of the $(l-1)$-th decoder layer, the output of the last encoder layer and the contextual grapheme representations of GBERT. 
Specifically, in each encoder layer,  an additional GBERT-Enc attention module is added to the original Transformer encoder layer in order to adaptively control how this layer  interacts with the  GBERT  representations. In each decoder layer, there is a similar GBERT-Dec attention module for a similar purpose. Moreover, a drop net is  used to regularize the network training, outputting one  of the two inputs or their average results during training and outputting the average results during inference. More details can be found in the original paper of the BERT-fused model \cite{zhu2019incorporating}.

	\begin{table}[tp]
	
	\centering
	\caption{The four languages used in our experiments.}  
	\label{tab:language}

			\resizebox{\linewidth}{!}{	\begin{tabular}{cccc}
				\hline
				\textbf{Language} & \textbf{Language Family} & \textbf{Script Type} & \textbf{Word Example}\\
				\hline 
				Dutch & Germanic & Latin &
				afnemer \\
				Serbo-Croatian & South Slavic &
				Latin & bu\dj enje \\
				Bulgarian & East Slavic & 
				Cyrillic & абоната \\ 
				Korean & Koreanic & Hangul
				& 가치관\\
				\hline
	\end{tabular}
	}
\end{table}
				
	\section{Experiments}
\label{sec:experiments}
\subsection{Datasets}
\label{ssec:subhead}
Experiments were conducted on the datasets of the  SIGMORPHON 2021 G2P task \cite{ashby2021results}. 
The four most difficult languages (i.e., the languages with the highest word error rates) among all ten languages in the medium-resource subtask of the SIGMORPHON 2021 task was chosen, according to the G2P performance of the official baseline \cite{clematide2021cluzh}.
As listed in Table~\ref{tab:language}, they belong to different language families and have different script types.
In the official settings, 8000, 1000 and 1000 pronunciation records were used for training, validation and test for each language. Following Elsaadany and Suter \cite{elsaadany2020grapheme},  Korean letters were decomposed into single-sound letters with \textsf{hangul-jamo}\footnote{\url{https://github.com/jonghwanhyeon/hangul-jamo}}, e.g., 가감  $\rightarrow$ ㄱ ㅏ ㄱ ㅏ ㅁ.


A \emph{medium-resource} G2P task and a \emph{low-resource}  G2P task were designed based on the datasets of these four languages. For the medium-resource task, the training, validation and test sets were the same as the official settings.
For the low-resource task, we randomly sampled 1000  records from the original training set  to form a smaller training set for each language, and used the original validation and test sets. 
Thus, the test set performance of these two tasks can be compared directly.

In addition, the word list for each language was collected from WikiPron \cite{lee2020massively} for GBERT pre-training. 
The words in the validation and test sets of the G2P tasks were excluded and the final word lists for these four languages in Table \ref{tab:language} contained 27.0k, 35.0k, 43.1k, and 14.1k words  respectively.
For each language,  90\% of the words were used  to train the monolingual GBERT and the remaining 10\%  were used for validation. 

		\begin{table*}[tp]\footnotesize
		
		\centering
		\caption{The WER and PER results (\%) of different models on our medium-resource and low-resource G2P tasks. 
		}  
		\label{tab:tab2}
\resizebox{0.85\linewidth}{!}{
			\setlength{\tabcolsep}{1mm}{
				\begin{tabular}{ccccccccc}
					\hline
					\multirow{2}*{Model} & 
					\multicolumn{2}{c}{Dutch}
					& \multicolumn{2}{c}{Serbo-Croatian}  & \multicolumn{2}{c}{Bulgarian} & \multicolumn{2}{c}{Korean}  \\
					& WER & PER & WER & PER & WER & PER & WER & PER \\
					\hline 

					\emph{medium-resource}  & & & & & & \\
                  \hline
                IL \cite{clematide2021cluzh} 
                       & 17.7  ± 1.3 & -   &  38.9  ±1.2 & - & 19.7  ± 1.7 & - & 18.9  ± 0.8 & - \\
					Transformer  &16.86 ± 0.14 & 3.49 ± 0.05  &  
					39.50 ± 0.71 & 7.94 ± 0.14  &  20.72 ± 2.70 & 3.84 ± 0.74   
					& 
					19.26 ± 0.48 & 3.39 ± 0.10
					\\
			    				GBERT w/o fine-tuning    &19.98 ± 0.66 & 4.61 ± 0.48 &  43.92 ± 0.97 & 9.53 ± 0.75   & 23.24 ± 2.54 & 4.17 ± 0.40   &
				23.06 ± 0.30 & 4.92 ± 0.15  \\
				   GBERT fine-tuning & 16.18 ± 0.23 & 3.59 ± 0.39 & 39.08 ± 1.05 &  7.83 ± 0.21 & 18.88 ± 2.31 & \bf 3.42 ± 0.33 &  19.80 ± 0.62 & 3.52 ± 0.12 \\
				 GBERT attention   & \bf 15.86 ± 0.21 &  \bf 3.38 ± 0.07 &
				 \bf 37.64 ± 0.71 & \bf7.67 ± 0.20 &  \bf 18.60 ± 1.92 & \bf 3.42 ± 0.34 
				 &
				\bf 17.94 ± 0.71 &
				\bf 3.16 ± 0.13
				  \\
				
				
					\hline
					

				\emph{low-resource} & & & & & & & & \\
                \hline
					Transformer & 34.30 ± 0.66 & 9.39 ± 0.42 &  68.86 ± 1.08 & 15.41 ± 0.16  &  33.06 ± 1.90 &  6.07 ± 0.35  & \bf 29.72 ± 0.77 & \bf 6.79 ± 0.35   \\
					GBERT w/o fine-tuning & 34.56 ± 0.72 & 9.13 ± 0.60   & 69.70 ± 0.69 & 15.88 ± 0.65  & 41.87 ± 0.89 & 9.15 ± 0.88  &  42.76 ± 1.04 & 12.66 ± 0.29 \\
						GBERT fine-tuning &  \bf 28.96 ± 0.69 & \bf 6.94 ± 0.61 &  \bf 63.12 ± 0.76 & \bf 13.29 ± 0.57 & \bf 30.86 ± 1.66 & \bf 5.30 ± 0.36  & 32.78 ± 1.21 & 8.43 ± 0.11 \\
						GBERT attention  &  35.28 ± 0.66 & 9.41 ± 0.33 & 68.14 ± 0.68 & 15.06 ± 0.29  &  31.98 ± 1.32 &  5.76 ± 0.35    &29.78 ± 0.60 & 6.81 ± 0.30    \\		
					\hline	

		\end{tabular}}}
	\end{table*}

\subsection{Implementation}
\label{ssec:implementation}
The  GBERT for each language was a 6-layer Transformer encoder. 
Other hyperparameters of GBERT followed previous work    
\cite{elsaadany2020grapheme}. 
The ratio of masked graphemes  in a word was set as 20\% for pre-training GBERT, which was higher than the ratio of masked words (15\%)  in  BERT since we expected  enough masked graphemes in short words. Our source code is released \footnote{\url{https://github.com/ldong1111/GraphemeBERT}}.
Evaluating the pre-trained GBERT models on the validation set of each language, the prediction accuracies of the masked graphemes were 53.48\%, 58.43\%, 80.66\% and 40.63\%, respectively. We can observe that they were all much higher than the accuracy of random prediction ($\sim$ 2\%), which shows the  contextual relationship among graphemes in a word. 

Finally, we compared five models as follows.
\begin{itemize}
\item \textbf{Imitation learning (IL)  \cite{clematide2021cluzh}} This baseline model adopted  a neural transducer \cite{aharoni2017morphological} that operated over explicit edit actions with imitation learning, and achieved similar performance to the Transformer model in the SIGMORPHON 2020 G2P task \cite{makarov2020cluzh, elsaadany2020grapheme}. 
We did not reimplement this model and just reported the single-model results in the  paper of IL model \cite{clematide2021cluzh} on our medium-resource G2P task. 
\item \textbf{Transformer} This baseline model adopted a Transformer-based architecture \cite{vaswani2017attention} and 
was implemented by us following  previous work   \cite{elsaadany2020grapheme}. The model hyperparameters  were tuned for different tasks of different languages, according to the performance on the validation set.

\item  \textbf{GBERT w/o fine-tuning} This model followed the proposed method in Section \ref{ssec:finetuningbert} except that the parameters of GBERT were frozen when tuning other model parameters. 

\item \textbf{GBERT fine-tuning} This model followed the proposed method in Section \ref{ssec:finetuningbert}.
Like  previous work \cite{liu2019text},  different learning rates were used for the pre-trained GBERT encoder and the randomly initialized Transformer decoder during fine-tuning. Both learning rates were   selected from 1e-3, 5e-4, 3e-4, 1e-4, 1e-5 based on the performance of the validation set for different tasks.

\item \textbf{GBERT attention} 
This model followed the proposed method in Section \ref{ssec:bertattention}
and had the same hyperparameters as the baseline Transformer model. 
\end{itemize}

\subsection{Evaluation Metrics}
\label{ssec:metrics}
Word error rate (WER) and phoneme error rate (PER) were used as the evaluation metrics in our experiments.
WER is the percentage of words whose predicted phoneme sequences were not identical to the gold references.  PER is the sum of the Levenshtein distance between the predicted and the reference phoneme sequences, divided by the sum of the reference lengths on the test set.  The lower the WER and PER, the better the performance. 

We conducted each experiment  five times with different random seeds and reported the mean and standard  deviation of the five results for each model. Note that the results of the IL model were quoted and  10 repetitions were used by its authors \cite{clematide2021cluzh}.
Thus, its standard deviations can not be compared with other models.

\subsection{Results}
\label{ssec:results}



Table $\ref{tab:tab2}$ shows the  WER and PER results of different models on our medium-resource and low-resource G2P tasks. 
The Transformer baseline implemented by us outperformed the IL baseline on the medium-resource task of Dutch, but was not as good as IL on the other three languages.
For the GBERT w/o fine-tuning model, its performance was much worse than the Transformer baseline in almost all  experiments, especially for Korean. 
This indicates that the grapheme representations derived from the pre-trained GBERT can not provide all the necessary information for G2P conversion.  

%
%

On the medium-resource G2P task,  the GBERT fine-tuning model significantly outperformed its counterpart without fine-tuning for all four languages, and achieved lower WERs and PERs than the Transformer baseline for all languages except Korean.
The GBERT attention model obtained the lowest WERs and PERs among all five models for all languages. 
All these results demonstrate the effectiveness of our proposed methods of fine-tuning GBERT or fusing GBERT into Transformer for the medium-resource G2P task. 
Although the GBERT attention model also adopted GBERT as a feature extractor, just like the 
GBERT w/o fine-tuning model, its  advantage is that not only GBERT representations but also the original input of the Transformer-based G2P model is utilized by attention modules.

On the low-resource G2P task,  the performance of all models decreased significantly compared to the medium-resource scenario, indicating the importance of training data amount to the state-of-the-art neural G2P models. The GBERT fine-tuning model also outperformed the 
GBERT w/o fine-tuning model for all four languages and outperformed the Transformer baseline for all languages except Korean. 
However, the  GBERT attention model  performed  better than the Transformer baseline only for Serbo-Croatian and Bulgarian, and performed not as good as the GBERT fine-tuning model for Dutch, Serbo-Croatian and Bulgarian.
These results show that the method of fusing GBERT into Transformer may be more sensitive to the amount of training data than the method of fine-tuning GBERT because the former has a more complex model structure.
The reason that our proposed method did not achieve satisfactory performance for Korean on the low-resource G2P task may be that the mask prediction accuracy of Korean was much lower than that of the other three languages as shown in Section \ref{ssec:implementation}, i.e., the contextual relationship among graphemes in Korean was weaker  than that in the other three languages.

		\begin{table}[tp]
	\centering
	\caption{The influence of the masked grapheme ratio for pre-training GBERT on the performance of the GBERT attention model for the medium-resource Dutch G2P task.}  
	\label{tab:tab3}
	\resizebox{\linewidth}{!}{
		\begin{tabular}{cccc}
			\hline
			Mask Ratio  (\%)  & Mask Accuracy (\%) & WER (\%) & PER (\%)\\
			\hline
			15 & 53.01 & 16.34 ± 0.43 & 3.51 ± 0.08\\
			20 & 53.48 & 15.86 ± 0.21 & 3.38 ± 0.07 \\
			30 & 51.32 & 15.88 ± 0.30 & 3.40 ± 0.08  \\
			\hline
	\end{tabular}}
\end{table}

An additional experiment was conducted to investigate the influence  of the masked grapheme ratio for pre-training GBERT on the performance of our proposed methods.
The GBERT attention model for the medium-resource Dutch G2P task was taken as an example, and the results are shown in Table \ref{tab:tab3}.
We can observe that the higher mask ratios (20\% and 30\%) achieved lower WER and PER than the original ratio of 15\% in BERT. 
One possible reason is that the average number of graphemes in a word is much smaller than the average number of words in a sentence and the grapheme space is much more compact than the word space. Thus, low mask ratios may constrain the model from discovering the contextual relationship among graphemes. 

\subsection{GBERT-based Transfer Learning For G2P Conversion}
\label{ssec:multilingual_experiment}
We also conducted an experiment to explore GBERT-based transfer learning, where the G2P data of another high-resource language was used to improve the performance of the low-resource G2P task.
In this experiment, Dutch was selected as the low-resource language and English was chosen as the high-resource language since they belong to the same Germanic language family and have the same Latin script type.
The supervised training data of English contained 33,344 word-pronunciation pairs from the high-resource subtask of the SIGMORPHON 2021 task.

First, a bilingual GBERT was pre-trained by mixing 49.1k English words from WikiPron and the Dutch word list mentioned in Section \ref{ssec:subhead}.
Following a variant of BERT \cite{song2019mass}, we added a language embedding to the input tokens of GBERT to distinguish different languages.  
Then, neural G2P models were built using the supervised training data of both English and Dutch. Following previous work \cite{elsaadany2020grapheme}, the words prefixed with a language tag were used as model inputs. The results of different models are shown in Table $\ref{tab:tab4}$.

\begin{table}[tp]
	
	\centering
	\caption{The WER and PER results (\%) of different models by transfer learning with high-resource English G2P data on the low-resource Dutch G2P task.
	}  
	\label{tab:tab4}

	\resizebox{0.7\linewidth}{!}{
		\setlength{\tabcolsep}{1mm}{
			\begin{tabular}{ccc}
				\hline
				Model & WER & PER
				\\
				
				\hline
				Transformer  &24.54 ± 0.53 & 5.09 ± 0.22  
				\\
				GBERT w/o fine-tuning    & 33.30 ± 0.89 & 7.89 ± 0.25    \\
				GBERT fine-tuning & 23.42 ± 0.82 & 4.84 ± 0.16 \\
				GBERT attention   &  \bf 23.36 ± 0.82 & \bf 4.76 ± 0.25 
				\\
				
				\hline
	\end{tabular}}}
\end{table}


Comparing  Table \ref{tab:tab4} with the low-resource results in Table  \ref{tab:tab2},
we can see that the WERs and PERs of all systems degraded significantly after transfer learning. Different from the  low-resource results in Table  \ref{tab:tab2}, the GBERT attention model achieved the lowest WER and PER among all four models. This indicates that the GBERT attention model can benefit from the increased training data from similar high-resource languages, and our proposed GBERT-based methods can also be combined with the transfer learning strategy to obtain better performance.

\section{Conclusion}
\label{sec:pagestyle}
In this paper, we have proposed a  pre-trained grapheme  model GBERT that outputs contextual grapheme representations.  The training of  GBERT requires only easily accessible word lists.
In addition,  two methods, namely fine-tuning GBERT and fusing GBERT into Transformer, have been developed to enhance the state-of-the-art Transformer-based G2P model with GBERT.  Experiments on   the  Dutch, Serbo-Croatian, Bulgarian  and Korean datasets of the SIGMORPHON 2021 G2P task show that the method of fusing GBERT can reduce the WER and PER  of all languages on  the medium-resource G2P task, while the method of fine-tuning GBERT can improve the G2P performance of most languages  on the low-resource G2P task.
In the future, we intend to improve the pre-trained grapheme model by developing new architectures and loss functions, and utilizing phrase-level or sentence-level training data.

\bibliographystyle{IEEEbib}
\bibliography{strings,refs}

\end{document}